\PassOptionsToPackage{table}{xcolor}
\documentclass{fairmeta}

\usepackage{newpxtext,newpxmath}
\definecolor{hbblue}{HTML}{0668E1}
\colorlet{metablue}{hbblue}
\newcommand{\hbhl}[1]{\textcolor{hbblue}{\textbf{#1}}}

\usepackage{amsmath,amsfonts,bm}

\def\eqref#1{equation~\ref{#1}}

\def\1{\bm{1}}

\DeclareMathAlphabet{\mathsfit}{\encodingdefault}{\sfdefault}{m}{sl}
\SetMathAlphabet{\mathsfit}{bold}{\encodingdefault}{\sfdefault}{bx}{n}

\newcommand{\humanwinVideoReward}{43.3}
\newcommand{\humanwinQTwo}{55.4}
\newcommand{\humanwinVRInit}{51.2}

\usepackage{url}
\usepackage{pgfplots}
\pgfplotsset{compat=1.17}
\usetikzlibrary{arrows.meta,fit,positioning,tikzmark}
\usepackage{xspace}
\usepackage{enumitem}
\usepackage{wrapfig}
\usepackage{float}
\usepackage{environ}

\definecolor{tblhi}{HTML}{2166AC}
\definecolor{tbllo}{HTML}{B2182B}

\providecommand{\myfontsize}{\small}
\newcommand{\mytextbox}[2]{%
  \tikzmarknode[draw=#1,thick,inner sep=1pt]{test}{\myfontsize #2}%
}
\definecolor{reducedstepscolor}{rgb}{0.35, 0.35, 0.35}
\definecolor{weakgen13color}{rgb}{0.25, 0.45, 0.70}
\definecolor{weakgen5color}{rgb}{0.15, 0.30, 0.60}
\definecolor{alignedweakgencolor}{rgb}{0.40, 0.25, 0.65}
\definecolor{motionregencolor}{rgb}{0.80, 0.40, 0.10}
\definecolor{frameholdcolor}{rgb}{0.70, 0.20, 0.15}
\definecolor{temporalshufflecolor}{rgb}{0.10, 0.55, 0.50}
\definecolor{attributeeditcolor}{rgb}{0.65, 0.20, 0.50}

\newcommand{\reducedstepsname}{Reduced Steps\xspace}
\newcommand{\weakgenonename}{Weakest Gen (1.3B)\xspace}
\newcommand{\weakgenfivename}{Weak Gen (5B)\xspace}
\newcommand{\alignedweakgenname}{Aligned Weak Gen\xspace}
\newcommand{\motionregenname}{Motion Re-gen\xspace}
\newcommand{\frameholdname}{Frame Hold\xspace}
\newcommand{\temporalshufflename}{Temporal Shuffle\xspace}
\newcommand{\attributeeditname}{Attribute Edit\xspace}
\newcommand{\familyhead}[1]{\rotatebox{70}{\tiny\textbf{#1}}}

\newcommand{\reducedsteps}{\mytextbox{reducedstepscolor}{\textbf{\textcolor{reducedstepscolor}{\reducedstepsname}}}}
\newcommand{\weakgenone}{\mytextbox{weakgen13color}{\textbf{\textcolor{weakgen13color}{\weakgenonename}}}}
\newcommand{\weakgenfive}{\mytextbox{weakgen5color}{\textbf{\textcolor{weakgen5color}{\weakgenfivename}}}}
\newcommand{\alignedweakgen}{\mytextbox{alignedweakgencolor}{\textbf{\textcolor{alignedweakgencolor}{\alignedweakgenname}}}}
\newcommand{\motionregen}{\mytextbox{motionregencolor}{\textbf{\textcolor{motionregencolor}{\motionregenname}}}}
\newcommand{\framehold}{\mytextbox{frameholdcolor}{\textbf{\textcolor{frameholdcolor}{\frameholdname}}}}
\newcommand{\temporalshuffle}{\mytextbox{temporalshufflecolor}{\textbf{\textcolor{temporalshufflecolor}{\temporalshufflename}}}}
\newcommand{\attributeedit}{\mytextbox{attributeeditcolor}{\textbf{\textcolor{attributeeditcolor}{\attributeeditname}}}}

\newcommand{\bench}{\hbhl{\textsc{IsoVRBench}}\xspace}
\newcommand{\isovrm}{\textsc{IsoVRM}\xspace}
\newcommand{\isowan}{\textsc{IsoWan}\xspace}

\newcommand{\authornote}[3]{}

\newcommand{\mg}[1]{}
\newcommand{\jh}[1]{}
\newcommand{\karen}[1]{}
\newcommand{\stella}[1]{}
\newcommand{\han}[1]{}

\let\fairmetaabstract\abstract
\RenewEnviron{abstract}{\expandafter\fairmetaabstract\expandafter{\BODY}\global\let\abstractlist\abstractlist}

\title{IsoVRBench: Video Reward Modeling\\
Without Human Feedback}

\author[1,2]{Shuyue Stella Li}
\author[1,2]{Jacqueline He}
\author[2]{Jiahui Chen}
\author[2]{Sreya Dutta Roy}
\author[1]{Yulia Tsvetkov}
\author[1,2]{Luke Zettlemoyer}
\author[2,\dagger]{Marjan Ghazvininejad}
\author[1,2,\dagger]{Xiaochuang Han}

\affiliation[1]{University of Washington}
\affiliation[2]{Meta Superintelligence Labs}
\contribution[\dagger]{Equal last author}

\begin{abstract}

Video reward models are trained and evaluated on human preferences between videos that often differ \emph{simultaneously} on properties such as prompt alignment, visual fidelity, motion, and generator identity.
As a result, aggregate accuracy does not reveal which video properties a reward model can reliably distinguish.
We construct preference pairs by \hbhl{controlled interventions} that change one property of a generated video while holding the rest fixed. The resulting pairs, by construction, provide preferences for \emph{how} the videos differ along a specific dimension, without requiring any human annotation.
We propose \bench{}, a video generation reward modeling benchmark of $8{,}000$ such pairs across eight interventions, and show that state-of-the-art models separate broad quality differences between generators but miss important features such as temporal order.
The best reward model reaches only $60.8\%$ on reordered frames against $100\%$ for humans, and rankings on existing benchmarks do not predict this failure.
Using the synthetic isolated intervation pairs as \hbhl{training data} leads to 1) \hbhl{\isovrm}, a reward model outperforming its counterpart trained with a larger human labeled preference dataset by 3.8 points on VBench-2.0, 2) \hbhl{\isowan}, the video generator post-trained with \isovrm, outperforming its human-data counterpart in human preferences, and 3) gains that stack when combined with human data.
Because intervention pairs are automatically generated rather than collected, the benchmark and training data can be easily extended as generators evolve.
\end{abstract}

\date{\today}
\correspondence{\email{stelli@cs.washington.edu}}
\metadata[Code]{\url{https://github.com/stellalisy/IsoVRBench}}
\metadata[Data]{\url{https://huggingface.co/datasets/stellalisy/IsoVRBench}}

\begin{document}

\maketitle

\section{Introduction}
Reinforcement learning post-trains video generators against reward models~\citep{videoreward,dpo,diffusiondpo}, which rank candidates at sampling time and define the objective for preference optimization.
A generator therefore improves only along the properties its reward model perceives.
Yet video reward models (VRMs) are trained and evaluated on human preferences between independently generated videos, which differ simultaneously in visual quality, content, motion, and generator identity.
Neither the preference labels nor a model's accuracy on these comparisons reveals which properties the model perceives, even when annotators are instructed to judge a particular dimension.
Existing benchmarks also rely on outputs from fixed generator sets and per-sample human labels~\citep{videoreward,genaibench,mjvideo,vbench2}, which limits their scale and freshness as generators and their failure modes evolve.

\begin{figure}[t]
\centering
\includegraphics[width=\linewidth]{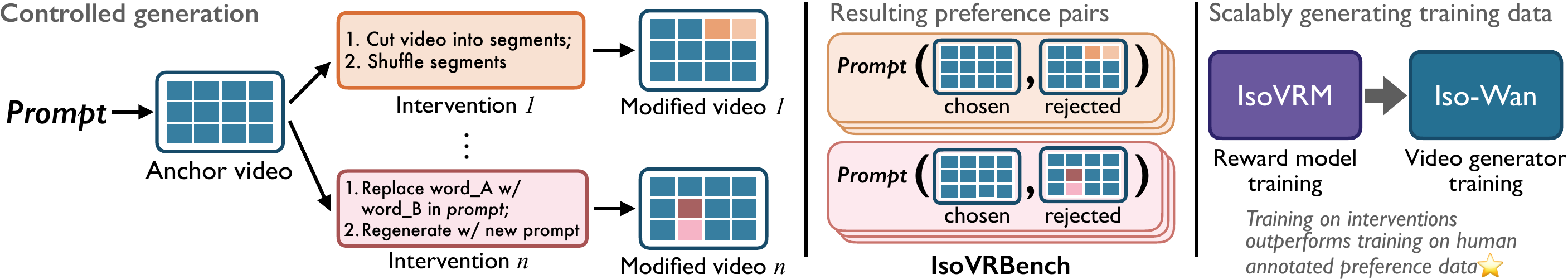}\vspace{-2mm}
\caption{\textbf{Controlled intervention data generation procedure.}
A prompt is rendered once to create an anchor video.
A controlled intervention (e.g., shuffling segments of a video) automatically produces a modified counterpart with a predetermined preference label.
The resulting pairs serve reward model evaluation (\bench{}), reward model training, and video generator post-training.
}\vspace{-2mm}
\label{fig:teaser}
\end{figure}

We construct comparisons by \hbhl{controlled intervention} instead, and use them to evaluate reward models, train them, and then post-train the video generators (Figure~\ref{fig:teaser}).
Starting from an anchor video, an intervention changes one property while holding the rest fixed, so both the preferred side and the property under test follow from the construction.
\temporalshuffle{}, for example, preserves every frame of the anchor but changes their order, testing whether a model uses frame order, while \framehold{} creates stuttering motion from the anchor's own frames.
No pair needs a human label, and human validation is needed only once per intervention to confirm that the intended change is perceptible.
Because the same interventions apply to any generator's outputs, the pairs can be regenerated as generators evolve.
We use this procedure to construct \bench{}, our benchmark of $8{,}000$ such pairs from eight interventions that target motion, text alignment, visual quality, and overall quality.

Evaluating nine VRMs and two general-purpose vision-language judges on \bench{}, we find that models are substantially more accurate on broad differences between generators than specific and easily perceivable differences such as temporal changes that preserve visual content.
On shuffled videos, the best reward model reaches only $60.8\%$ against $100\%$ for humans, and no model exceeds $66.1\%$. %
Rankings on existing reward and temporal-understanding benchmarks do not predict sensitivity to frame order.

The same construction yields a training corpus of $160{,}000$ pairs.
\hbhl{\isovrm-Q2}, trained on these intervention pairs, outperforms its counterpart VideoReward \citep{videoreward}, which is trained from the same base model on a larger human-annotated dataset by $3.8$ points on VBench-2.0.
When used as the reward for post-training a video generator, \isovrm-Q2 yields the generator preferred by human raters over the generator trained with VideoReward.
\hbhl{\isovrm-VR}, initialized from VideoReward and further trained on intervention pairs, outperforms VideoReward, including $63.5\%$ versus $56.3\%$ on VBench-2.0 and $69.0\%$ versus $66.4\%$ on MJ-Bench-Video. %
The generator post-trained with \isovrm-VR reaches the highest VBench-2.0 score, $51.5$ against $49.0$ under VideoReward, with a human win frequency of $\humanwinVRInit\%$.
Thus, intervention-derived supervision can either replace human preference supervision or augment it, improving both reward-model transfer and the generators trained against those rewards.
We make three contributions.\vspace{-1mm}

\begin{itemize}[itemsep=1pt,topsep=2pt,leftmargin=12pt]
  \item \hbhl{Controlled intervention pipeline for scalable VRM development}, where changing one property at a time determines both the preference label and the property under test without human labeling.
  \item \textbf{\bench{}}, a benchmark on which current VRMs and general-purpose judges perceive frame content but not frame order, a failure that rankings on existing benchmarks do not predict.
  \item \hbhl{Intervention-derived supervision} that trains competitive VRMs without human preference labels or in addition to them, and improves downstream video-generator post-training.
\end{itemize}

\section{Constructing \bench{}}

\label{sec:method}

\subsection{Controlled intervention framework}

Let $x$ be a video generated from prompt $p$ under sampling configuration $\theta$.
An intervention $\mathcal{I}_D$ produces a counterpart $x'$ that differs from the \emph{anchor} $x$ along a set of dimensions $D$, either by transforming $x$ itself or by rendering a new video after changing $p$ or $\theta$.
Each intervention targets a set of dimensions $D$ for which the construction specifies the preferred side; for the interventions in this work, the anchor $x$ is preferred, such that $x \succ_d x'$ for each $d \in D$.
For an intervention to provide a controlled evaluation of a particular property, we consider three conditions.

\begin{enumerate}[itemsep=0pt,topsep=0pt,leftmargin=15pt]
  \item \textbf{Perceptibility.}
  The induced change should be detectable to human viewers, who should prefer the intended side on the targeted dimension.
  \item \textbf{Confinement.}
  The intervention should change \emph{only} the targeted dimension, while holding other dimensions fixed. This enables differences in reward to be attributed to the target property.
  \item \textbf{Plausibility.}
  The resulting counterpart must resemble a plausible generator output rather than an easily exploitable artifact unrelated to the targeted property.
\end{enumerate}

Section~\ref{sec:validation} validates perceptibility through human judgments, and Appendix~\ref{app:validation} provides additional validation of confinement.

\subsection{Intervention definitions}
\label{sec:interventions}
\begin{figure}[t]\vspace{-1mm}
\centering
    \centering
    \includegraphics[width=0.85\linewidth]{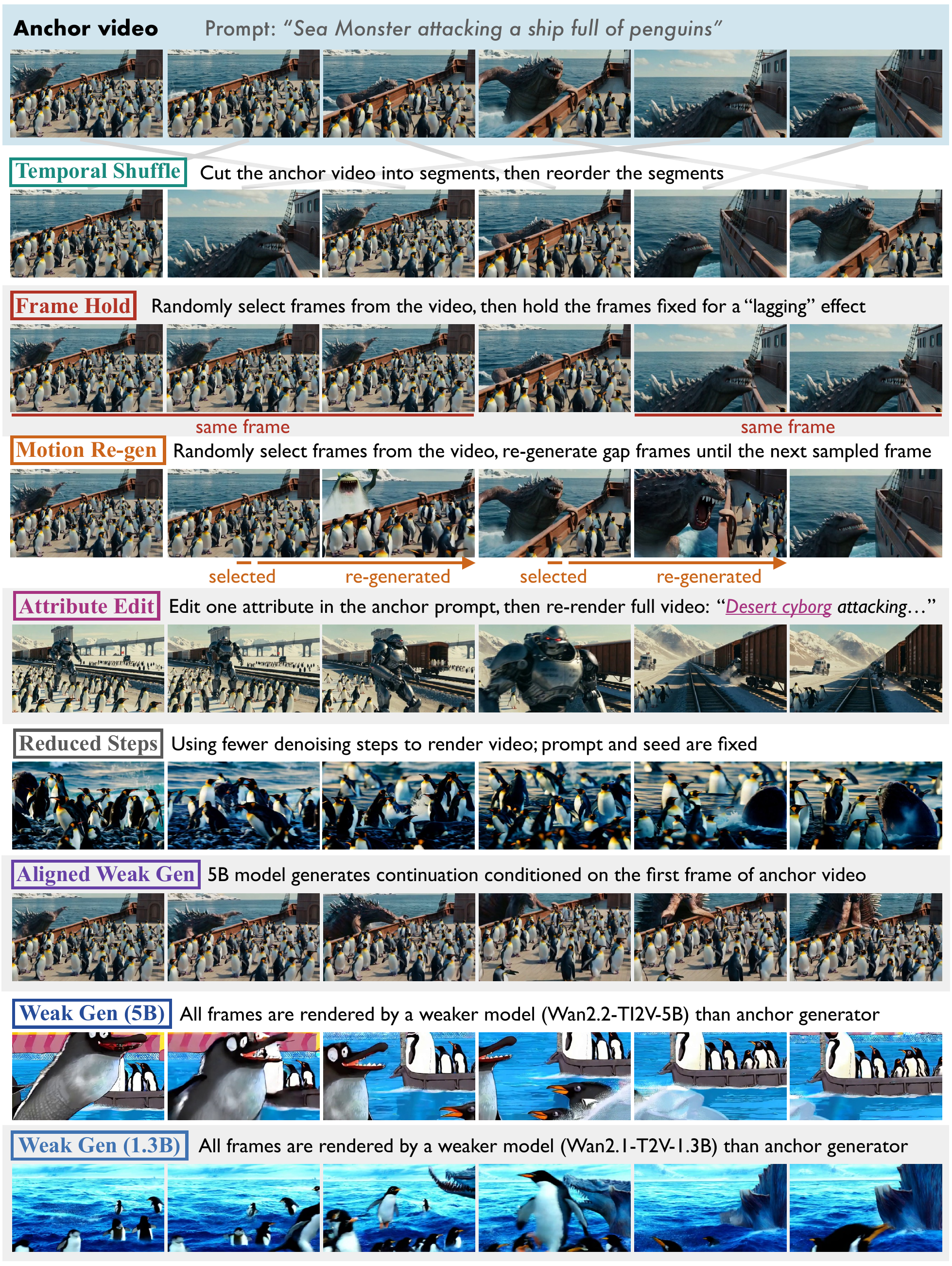}\vspace{-3mm}
\caption{\textbf{\bench{} interventions on a representative prompt.}
The top row is the anchor prompt and every row below is a corresponding clip generated from each intervention.
See Table \ref{tab:families} for intervention descriptions.
}\vspace{-4mm}
\label{fig:qualitative}
\end{figure}

\begin{table}[t]
\centering
\small
\caption{\textbf{Interventions in \bench{}, with their descriptions and evaluation dimensions.}
\bench{} draws $1{,}000$ pairs from each intervention, resulting in $8{,}000$ total samples.}\vspace{-2mm}
\label{tab:families}
\renewcommand{\arraystretch}{1.15}
\begin{tabular}{lll}
\toprule
\textbf{Intervention} & \textbf{Description} & \textbf{Dimension} \\
\midrule
\multicolumn{3}{l}{\emph{Rearranging the anchor's frames}} \\
\temporalshuffle & Anchor cut into six segments and randomly reordered & Motion \\
\framehold       & Frames of the anchor held, giving stuttering motion & Motion \\
\midrule
\multicolumn{3}{l}{\emph{Regenerating one part of the clip}} \\
\motionregen     & Regenerated motion between real keyframes & Motion \\
\attributeedit   & Single-attribute change in input text prompt & Text \\
\reducedsteps    & 8 denoising steps (vs. 50 for the anchor) with fixed prompt & Visual \\
\midrule
\multicolumn{3}{l}{\emph{Substituting the generator}} \\
\alignedweakgen  & Wan2.2-TI2V-5B continuing the anchor's first frame & Overall \\
\weakgenfive     & Rendered by Wan2.2-TI2V-5B from fixed prompt & Overall \\
\weakgenone      & Rendered by Wan2.1-T2V-1.3B from fixed prompt & Overall \\
\bottomrule
\end{tabular}
\end{table}
In this work, we employ eight distinct interventions to construct video pairs along different quality dimensions.
Table~\ref{tab:families} summarizes the interventions by decreasing control over the anchor and Figure~\ref{fig:qualitative} shows one example of each intervention.

\textbf{Rearranging the anchor's frames.}
The most controlled interventions modify only the temporal presentation of frames from the anchor.
\temporalshuffle{} divides the anchor into six segments and permutes them so that no segment remains in place, changing the order of events while preserving every frame.
Because temporal order is informative only when there is sufficient motion, we use the $1{,}000$ anchors with the most motion (more details in Appendix~\ref{app:shuffle-motion}).
\framehold{} retains $8$ to $12$ anchor frames and repeats each until the next, preserving the clip length while turning smooth motion into visible jumps.
Because both interventions construct the counterpart entirely from anchor frames, they strongly control for frame-level appearance and are labeled on motion.
We call them the \hbhl{frame-reuse interventions}.

\textbf{Regenerating part of the clip.}
The next three interventions introduce newly generated content by changing a single input.
Together with the three below, they are the \hbhl{re-rendering interventions}, since their counterparts contain newly rendered frames.
\motionregen{} retains $8$ or $12$ anchor frames as keyframes and regenerates each interval using the anchor's own generator, starting from one keyframe and returning to the next.
This allows motion to deviate from the anchor's trajectory between fixed keyframes while preserving the surrounding scene.
Random keyframe spacing varies the duration over which this deviation can occur.
\attributeedit{} edits or removes one attribute of the prompt and re-renders using the same generator and seed; the resulting video is scored under the original prompt, making the edited attribute the targeted source of misalignment.
\reducedsteps{} re-renders the same prompt and seed using $8$ denoising steps instead of $50$ in the anchor.
The three interventions target motion, text alignment, and visual quality, respectively.

\textbf{Substituting the generator.}
The last three interventions replace the anchor's generator (Wan2.2-T2V-A14B) with a weaker generator while preserving the prompt.
\alignedweakgen{} renders with Wan2.2-TI2V-5B conditioned on the anchor's first frame, while \weakgenfive{} and \weakgenone{} render the prompt with Wan2.2-TI2V-5B and Wan2.1-T2V-1.3B, respectively.
Because changing the generator can simultaneously affect visual quality, motion and text alignment, these pairs receive an overall label and deliberately give up confinement. They resemble the independently generated comparisons used in existing preference benchmarks, allowing Section~\ref{sec:vgrb} to compare published rankings with performance on these interventions.

\subsection{Human validation of perceptibility}
\label{sec:validation}
We validate each intervention with human judgments of perceptibility.
For each intervention, raters compared $45$ intervention pairs and $8$ control pairs that show the same clip on both sides, with presentation order randomized and ties allowed.
Each pair was judged separately on visual quality, motion, and text alignment, and the \textbf{Human} row of Table~\ref{tab:main} reports agreement with the intervention label on its targeted dimension.
For the three generator substitutions, which carry an overall label, we define overall preference as the sign of the sum of the three dimension-level preferences, paralleling VideoReward's overall score as the sum of its three reward heads~\citep{videoreward}. %
Raters correctly identify all $64$ control pairs as ties.
Human agreement with the intervention label ranges from $98\%$ to $100\%$ on six of the eight interventions, with $93\%$ agreement on \framehold{} and $84\%$ on \reducedsteps{}.

\subsection{Corpus and splits}

\bench{} contains $1{,}000$ pairs per intervention.
In total, the benchmark contains $8{,}000$ pairs spanning $15{,}047$ clips and $7{,}129$ prompts, with chance accuracy at $50\%$.
For the intervention-strength analysis in Appendix~\ref{app:strength}, we additionally construct a \reducedsteps{} split using $16$ rather than $8$ denoising steps.
The training corpus contains $164{,}000$ pairs spanning $234{,}075$ clips and $70{,}075$ prompts, with $160{,}000$ pairs for training and $4{,}000$ for validation.
Prompts are disjoint across splits because multiple interventions can reuse the same anchor clip, making a pair-level split susceptible to video leakage. Benchmark prompts are also disjoint from the training corpus.

\section{Benchmarking reward models on \bench{}}
\label{sec:results}

We evaluate nine reward models: VideoReward~\citep{videoreward}, VisionReward~\citep{visionreward}, PickScore~\citep{pickscore}, UnifiedReward-Flex-9B and UnifiedReward-2.0-8B~\citep{unifiedreward}, VideoScore v1 and v1.1~\citep{videoscore}, VideoScore2~\citep{videoscore2}, and LiFT~\citep{lift}.
For models with property-specific reward heads, we use the head corresponding to the labeled property; for single-score models, we use their overall score.
To see whether weaknesses are specific to reward-model training, we also evaluate two general-purpose vision-language judges that are not trained as video reward models: Qwen3-VL-8B-Instruct~\citep{bai2025qwen3vltechnicalreport} and GPT-4.1~\citep{openai2025gpt41}.
Each judge scores a clip its prompt and eight uniformly sampled frames, using the expected value over the ten digit tokens, then compare the two scores from the pair. Evaluation details are in Appendix~\ref{app:judges}.

\begin{table}[t]
\caption{\textbf{Per-intervention accuracy (\%) on \bench, sorted in descending order by average performance.}
Ties are counted as incorrect; LiFT, VideoScore2, and VisionReward tie on $61\%$, $27\%$, and $21\%$ of pairs, and PickScore and VideoReward rarely tie (Appendix~\ref{app:tiehandling}).
Both UnifiedReward models also return no judgment on $78$ pairs, which count as wrong.
Accuracy over decided pairs, those on which a model prefers one side, is in Appendix~\ref{app:ties}.
}\vspace{-3mm}
\label{tab:main}
\centering\small\setlength{\tabcolsep}{4.5pt}
\resizebox{0.8\linewidth}{!}{
\begin{tabular}{lr|rrr|r|r|rrr}
\toprule
 & & \multicolumn{3}{c|}{\textbf{Motion}} & \textbf{Text} & \textbf{Visual} & \multicolumn{3}{c}{\textbf{Overall}} \\
\textbf{Model} & \textbf{Avg.} &
\familyhead{{\renewcommand{\myfontsize}{\tiny}\temporalshuffle{}}} &
\familyhead{{\renewcommand{\myfontsize}{\tiny}\framehold{}}} &

\familyhead{{\renewcommand{\myfontsize}{\tiny}\motionregen{}}} &

\familyhead{{\renewcommand{\myfontsize}{\tiny}\attributeedit{}}} &

\familyhead{{\renewcommand{\myfontsize}{\tiny}\reducedsteps{}}} &

\familyhead{{\renewcommand{\myfontsize}{\tiny}\alignedweakgen{}}} &

\familyhead{{\renewcommand{\myfontsize}{\tiny}\weakgenfive{}}} &

\familyhead{{\renewcommand{\myfontsize}{\tiny}\weakgenone{}}} \\
\midrule
VideoReward & 73.3 & \cellcolor{tblhi!14}\textbf{60.8} & \cellcolor{tblhi!28}\textbf{71.6} & \cellcolor{tblhi!45}\textbf{84.7} & \cellcolor{tblhi!31}74.1 & \cellcolor{tblhi!37}78.5 & \cellcolor{tblhi!31}73.6 & \cellcolor{tblhi!36}77.9 & \cellcolor{tblhi!19}64.9 \\
PickScore & 69.7 & \cellcolor{tbllo!6}45.0 & \cellcolor{tblhi!5}54.2 & \cellcolor{tblhi!24}68.3 & \cellcolor{tblhi!36}78.0 & \cellcolor{tblhi!23}68.0 & \cellcolor{tblhi!32}75.0 & \cellcolor{tblhi!51}\textbf{89.6} & \cellcolor{tblhi!38}79.4 \\
UnifiedReward-Flex-9B & 68.4 & \cellcolor{tbllo!3}47.7 & \cellcolor{tbllo!1}49.5 & \cellcolor{tblhi!6}54.3 & \cellcolor{tblhi!38}\textbf{78.9} & \cellcolor{tblhi!36}77.4 & \cellcolor{tblhi!25}69.2 & \cellcolor{tblhi!50}88.2 & \cellcolor{tblhi!41}\textbf{81.6} \\
UnifiedReward-2.0-8B & 68.1 & \cellcolor{tblhi!3}52.6 & \cellcolor{tblhi!6}54.3 & \cellcolor{tblhi!9}56.8 & \cellcolor{tblhi!28}71.7 & \cellcolor{tblhi!35}76.8 & \cellcolor{tblhi!22}66.7 & \cellcolor{tblhi!48}86.7 & \cellcolor{tblhi!38}79.4 \\
VideoScore v1 & 66.2 & \cellcolor{tbllo!1}49.5 & \cellcolor{tbllo!13}39.8 & \cellcolor{tblhi!33}75.2 & \cellcolor{tbllo!1}49.6 & \cellcolor{tblhi!40}\textbf{80.5} & \cellcolor{tblhi!32}74.4 & \cellcolor{tblhi!41}81.5 & \cellcolor{tblhi!37}78.8 \\
VideoScore v1.1 & 64.5 & \cellcolor{tblhi!5}54.2 & \cellcolor{tbllo!20}34.9 & \cellcolor{tblhi!19}64.7 & \cellcolor{tblhi!13}60.1 & \cellcolor{tblhi!26}70.2 & \cellcolor{tblhi!34}\textbf{76.3} & \cellcolor{tblhi!42}82.4 & \cellcolor{tblhi!30}72.9 \\
VisionReward & 58.0 & \cellcolor{tbllo!29}27.7 & \cellcolor{tbllo!24}31.7 & \cellcolor{tbllo!6}45.0 & \cellcolor{tblhi!21}66.2 & \cellcolor{tblhi!39}80.0 & \cellcolor{tblhi!6}54.8 & \cellcolor{tblhi!47}86.3 & \cellcolor{tblhi!29}72.6 \\
VideoScore2 & 47.9 & \cellcolor{tbllo!15}38.2 & \cellcolor{tbllo!18}35.9 & \cellcolor{tbllo!10}42.2 & \cellcolor{tblhi!16}62.1 & \cellcolor{tbllo!3}47.6 & \cellcolor{tbllo!4}47.0 & \cellcolor{tblhi!14}61.0 & \cellcolor{tbllo!1}49.4 \\
LiFT & 19.5 & \cellcolor{tbllo!50}11.9 & \cellcolor{tbllo!53}8.9 & \cellcolor{tbllo!42}18.0 & \cellcolor{tbllo!31}26.0 & \cellcolor{tbllo!35}22.8 & \cellcolor{tbllo!37}21.9 & \cellcolor{tbllo!37}21.5 & \cellcolor{tbllo!33}24.8 \\
\midrule
\multicolumn{10}{l}{\emph{general vision-language models prompted as judges}} \\
GPT-4.1 & 70.1 & \cellcolor{tblhi!12}58.9 & \cellcolor{tblhi!19}64.7 & \cellcolor{tblhi!10}57.5 & \cellcolor{tblhi!28}71.4 & \cellcolor{tblhi!35}77.1 & \cellcolor{tblhi!25}69.6 & \cellcolor{tblhi!46}85.5 & \cellcolor{tblhi!34}76.0 \\
Qwen3-VL-8B & 69.3 & \cellcolor{tblhi!9}57.1 & \cellcolor{tblhi!14}60.8 & \cellcolor{tblhi!19}64.8 & \cellcolor{tblhi!31}74.2 & \cellcolor{tblhi!31}73.5 & \cellcolor{tblhi!25}68.9 & \cellcolor{tblhi!42}82.6 & \cellcolor{tblhi!30}72.8 \\
\midrule
Human & 96.4 & \cellcolor{tblhi!65}100 & \cellcolor{tblhi!56}93 & \cellcolor{tblhi!65}100 & \cellcolor{tblhi!62}98 & \cellcolor{tblhi!44}84 & \cellcolor{tblhi!62}98 & \cellcolor{tblhi!62}98 & \cellcolor{tblhi!65}100 \\
\bottomrule
\end{tabular}
}
\end{table}

\textbf{Humans outperform every reward model on every intervention.}
Table~\ref{tab:main} reports performance on \bench{} on each isolated dimension.
The best reward model, VideoReward, averages $73.3\%$, and human raters average $96.4\%$, agreeing with the constructed label on $347$ of $360$ rated pairs (Appendix~\ref{app:gradient}).
The best model differs across interventions, and five of the nine reward models are best on at least one.

\textbf{Reward models fail when only the frame order changes.}
On every \hbhl{re-rendering intervention}, at least one reward model scores $76.3\%$ or higher, while on the two \hbhl{frame-reuse interventions} the best model, VideoReward, reaches $60.8\%$ on \temporalshuffle{} and $71.6\%$ on \framehold{}, against $100\%$ and $93\%$ for human raters.
Every reward model scores lower on both frame-reuse interventions than on its best re-rendering intervention, and six of the nine are at or below chance on \temporalshuffle{}.
Section~\ref{sec:framesweep} tests whether the number of frames a model sees explains this gap.

\textbf{General-purpose VLM judges share the failure.}
GPT-4.1 and Qwen3-VL-8B, which are not trained as video reward models, rank second and fourth of the eleven models, at $70.1\%$ and $69.3\%$.
They show the same contrast, scoring $85.5\%$ and $82.6\%$ on \weakgenfive{}, where a smaller generator renders the same prompt, but only $58.9\%$ and $57.1\%$ on \temporalshuffle{}, where the anchor's own frames are reordered.
The failure is therefore not specific to reward-model training.

\textbf{Average accuracy hides differences between models.}
The five models whose averages lie between $68.1\%$ and $70.1\%$ differ by up to $15.2$ points on a single intervention, most on the three motion interventions ($13.9$ on \temporalshuffle{}, $15.2$ on \framehold{}, and $14.0$ on \motionregen{}).
PickScore and UnifiedReward-Flex-9B average $69.7\%$ and $68.4\%$, yet PickScore scores $14.0$ points higher on \motionregen{} and $9.4$ points lower on \reducedsteps{}.
Fine-grained scores in each isolated dimension therefore allow us to better understand the gap in current models and enable targeted improvements.

\section{Analysis}
\label{sec:analysis}

Models struggle the most on the \hbhl{frame-reuse interventions}, specifically \temporalshuffle{} and \framehold{}, the two interventions where every frame is the anchor's own, so failures in these dimensions come from the models' inability to identify temporal relations and smoothness.
We show these failures are overlooked by existing benchmarks and are not recovered by feeding more frames into the context.

\subsection{No published benchmark predicts frame order sensitivity}
\label{sec:vgrb}

\begin{figure}[t]
    \centering
    \includegraphics[width=0.9\linewidth]{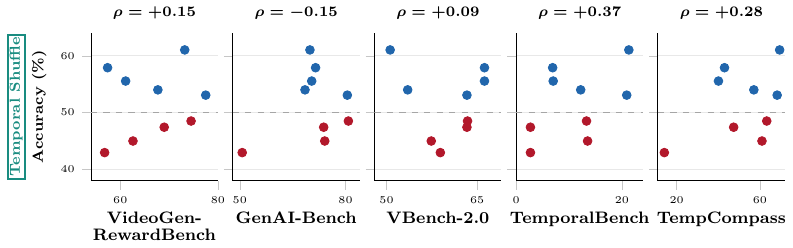}\vspace{-2mm}
    \caption{\textbf{Accuracy on each published benchmark against \bench{} \temporalshuffle{} accuracy.} Each reward model is a dot, and each horizontal axis uses the benchmark's reported metric.
    Blue is above chance on \temporalshuffle{}, red is below, and the dashed line is at chance. $\rho$ is the Spearman rank correlation down each panel, and none differs from zero under an exact permutation test.
    Details in Appendix~\ref{app:external}.
    }
\label{fig:temporal_shuffle}
\end{figure}

We compare model rankings on \temporalshuffle{} against two classes of benchmark, video generation reward benchmarks that compare independently generated videos---VideoGen-RewardBench~\citep{videoreward}, GenAI-Bench~\citep{genaibench}, and VBench-2.0~\citep{vbench2}---and temporal video-language understanding benchmarks---TemporalBench~\citep{temporalbench} and TempCompass~\citep{tempcompass})---which present a single video and evaluate its relationship to text.
Figure~\ref{fig:temporal_shuffle} shows the performance correlations between \temporalshuffle{} and each of the existing benchmarks.

\textbf{Rankings on existing benchmarks do not predict frame-order accuracy.}
Across the nine reward models, rank correlations between each published benchmark and \temporalshuffle{} range from $-0.15$ to $+0.37$, and none is significant (Figure~\ref{fig:temporal_shuffle}).
The two UnifiedReward models lead VideoGen-RewardBench and GenAI-Bench but score only $53.1\%$ and $48.5\%$ on \temporalshuffle{}; VideoReward, the decidedly best reward model on \temporalshuffle{} and \framehold{}, only ranks third, seventh, and last on the three video reward benchmarks.
On the two temporal-understanding benchmarks, VideoScore v1.1 and v1 score below random on both TemporalBench and TempCompass yet rank second and third among reward models on decided \temporalshuffle{} pairs, at $57.9\%$ and $55.6\%$.
These rankings therefore give no indication of whether a reward model perceives frame order.
Appendix~\ref{app:indep} finds the same with \bench{}'s own generator-substitution pairs, whose ranking does not predict the ranking on \temporalshuffle{} either.

\textbf{Existing benchmarks distinguish models only by generator identity and text alignment.}
We further compute the correlations between the existing benchmarks to each of the intervention families in \bench{} (Table~\ref{tab:rhofamily}).
We find that \textit{generator identity heavily dominates the preference labels in video reward model benchmarks}, where learning a predictor of generator identity, without seeing the video, can achieve higher performance than most state-of-the-art reward models (Appendix~\ref{app:generator-identity}).
GenAI-Bench ranks the nine models exactly as \weakgenfive{} does ($\rho=+1.00$) and nearly as \weakgenone{} does ($\rho=+0.87$), so its ranking coincides with how well a model separates a stronger generator from a weaker one, and the two generators' identities alone predict $74.1\%$ of its labels.
TempCompass, built to test temporal understanding, instead tracks \attributeedit{} ($\rho=+0.80$), the intervention that tests whether a video matches its prompt.
Neither kind of benchmark therefore isolates frame order, which \bench{} identifies as the prominent failure mode of current reward models.

\subsection{More frames do not improve temporal judgments}
\label{sec:framesweep}

\begin{figure}[t]
    \centering
    \includegraphics[width=0.95\linewidth]{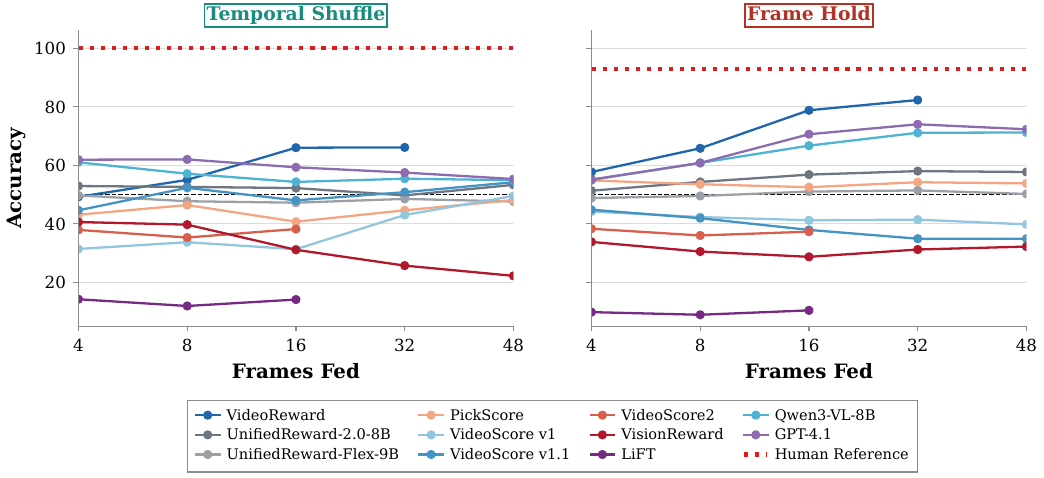}
    \caption{\textbf{Accuracy (\%) against the frame count on two interventions that reuse the anchor's frames.}
    Absent points mark frame counts a model cannot be scored at (Appendix~\ref{app:framesweep}).
    Seven of the eleven sit at or below chance at every frame count on at least one of the two interventions, and VisionReward falls on \temporalshuffle{} as it is given more frames.
    }
\label{fig:framesweep}
\end{figure}

The default frame count for each model is used in Table~\ref{tab:main}. We now vary the frame count for each model to investigate whether seeing more frames improves the temporal judgments of VRMs.
Figure~\ref{fig:framesweep} shows each checkpoint re-evaluated on the same $2{,}000$ pairs with $4$, $8$, $16$, $32$, and $48$ frames. %

\textbf{More frames help slightly on \framehold{} but not on \temporalshuffle{}.}
Averaged over the eleven models, moving from four frames to the largest frame count each model accepts raises accuracy by $4.4$ points on \framehold{} and $1.5$ points on \temporalshuffle{} (Figure~\ref{fig:framesweep}).
VideoReward, GPT-4.1, and Qwen3-VL-8B gain, respectively, $24.6$, $18.9$, and $16.2$ points on \framehold{} between four frames and their best frame count, with VideoReward reaching $82.3$.
On \temporalshuffle{}, only VideoReward improves substantially, by $16.9$ points to $66.1\%$, while GPT-4.1 and Qwen3-VL-8B lose $6.6$ and $6.1$ points, and VisionReward and LiFT stay below chance on both interventions at every frame count.
Seeing more frames therefore does not give these models access to frame order.
Denser sampling makes the repeated frames of \framehold{} easier to detect, consistent with VideoReward's gap between the two interventions widening from $8.5$ points at four frames to $16.2$ at thirty-two, whereas detecting a shuffle requires relating frames that are far apart in time.
Training on intervention pairs instead raises \temporalshuffle{} accuracy on held-out prompts to $84.9\%$ and $89.5\%$ (Appendix~\ref{app:ours-iso}).

\section{Intervention pairs as training data}
\label{sec:training}

\subsection{Training reward models}
We train two reward models on preference labels constructed by our intervention protocol.
Both employ a Qwen2-VL-2B backbone and follow the VideoReward architecture~\citep{videoreward}, with a scalar reward head at per-dimension marker tokens and a Bradley-Terry objective with ties~\citep{bradleyterry}, and Appendix~\ref{app:training-hparams} gives the training configuration.

We first ask whether intervention supervision can \emph{replace} human preference supervision by training \isovrm-Q2 from a Qwen2-VL-2B initialization without prior reward-model training.
We next ask whether intervention supervision can \emph{augment} human preference supervision, and train \isovrm-VR by initializing from released VideoReward weights, which have already been trained on human preference data.

Our intervention training data contains $20{,}000$ pairs per intervention, or $160{,}000$ in total, with another $500$ prompt-disjoint pairs per intervention for validation.
All \bench{} prompts are excluded, although the benchmark uses the same eight interventions (Appendix~\ref{app:ours-iso}).
We benchmark our reward models against VideoReward, which shares the same backbone, model size, and reward architecture but receives solely human preference supervision during reward-model training.

\begin{figure}[t]\vspace{-3mm}
\begin{minipage}{\textwidth}\vspace{3mm}
    \begin{minipage}{0.65\textwidth}
\captionof{table}{Transfer accuracy discarding ties (\%) on benchmarks built from human labels. \isovrm-VR, trained on intervention pairs on top of human labels, generally performs the best.
}\label{tab:transfer}
\vspace{-3mm}
\centering\small\setlength{\tabcolsep}{2.6pt}
\resizebox{\linewidth}{!}{
\begin{tabular}{lrrrr}
\toprule
\textbf{Model} & \textbf{VideoGen-RewardBench} & \textbf{GenAI-Bench} & \textbf{VBench-2.0} & \textbf{MJ-Bench-Video} \\
Benchmark Size & $20{,}997$ & $1{,}270$ & $11{,}029$ & $116$ \\
\midrule
\hbhl{\isovrm-Q2}      &  $71.5$ & $67.9$ & $60.1$ & $61.2$ \\
\hbhl{\isovrm-VR} & $\mathbf{73.3}$ & $68.2$ & $\mathbf{63.5}$ & $\mathbf{69.0}$  \\
\midrule
VideoReward      & $73.2$ & $\mathbf{69.9}$ & $56.3$ & $66.4$  \\
\bottomrule
\end{tabular}
}
    \end{minipage}
\hfill
    \begin{minipage}{0.33\textwidth}%
\captionof{table}{VBench-2.0 and 3-way human win-rate (\%) of trained video generators.%
}\label{tab:grpo}\vspace{-2.5mm}
\centering\small\setlength{\tabcolsep}{4pt}
\resizebox{\linewidth}{!}{
\begin{tabular}{lrr}
\toprule
\textbf{GRPO Reward} & \textbf{VBench-2.0} & \textbf{Human} \\
\midrule
Pretrained    & $47.3$ & \textemdash \\
\midrule
VideoReward & $49.0$ & $\humanwinVideoReward$ \\
\hbhl{\isovrm-Q2}     & $49.8$ & $\mathbf{\humanwinQTwo}$ \\
\hbhl{\isovrm-VR}     & $\mathbf{51.5}$ & $\humanwinVRInit$ \\
\bottomrule
\end{tabular}\vspace{-2mm}
}
    \end{minipage}%
\end{minipage}\vspace{2mm}
\end{figure}

As Table~\ref{tab:transfer} shows, \isovrm-Q2, trained without human preference labels, outperforms VideoReward on VBench-2.0 by $3.8$ points, $60.1$ against $56.3$, and trails it on VideoGen-RewardBench ($71.5$ against $73.2$), GenAI-Bench ($67.9$ against $69.9$), and MJ-Bench-Video ($61.2$ against $66.4$).
VBench-2.0 includes two generators absent from VideoReward's training data, and VideoReward rates both below their human win rates and the two it trained on above them (Appendix~\ref{app:shortcut-failure}).
This is the generator-identity shortcut of Section~\ref{sec:related}, since in preference benchmarks the two generators' identities alone predict $74.1\%$ to $78.2\%$ of the labels (Appendix~\ref{app:generator-identity}).
Intervention pairs carry little of this signal, because five of the eight interventions render both sides with the same generator, and those five supply most of the transfer.
Added to VideoReward's supervision, the intervention pairs raise VBench-2.0 from $56.3$ to $63.5$ in \isovrm-VR, which also scores $73.3$ against $73.2$ on VideoGen-RewardBench and $69.0$ against $66.4$ on MJ-Bench-Video~\citep{mjvideo}, and trails on GenAI-Bench, scoring $68.2$ against $69.9$.

\subsection{Training a video generator with intervention-supervised reward models}
\label{sec:generator-training}

We fine-tune Wan2.2-TI2V-5B with group relative policy optimization (GRPO) for diffusion models~\citep{liu2025flow} using each reward model in Table~\ref{tab:transfer}.
Appendix~\ref{app:grpo-hparams} shows more training details.

Human annotators prefer the generator post-trained under \isovrm-Q2, whose reward model saw no human preference labels.
Across every comparison in the three-way human study, it wins $266$ of its $480$ appearances, the \isovrm-VR generator wins $246$, and the VideoReward-supervised generator wins $208$, win frequencies of $\humanwinQTwo\%$, $\humanwinVRInit\%$, and $\humanwinVideoReward\%$ (Appendix~\ref{app:grpo-human} reports the pairwise rates and agreement).
On VBench-2.0 all runs improve on the pretrained generator (Table~\ref{tab:grpo}), and the generators post-trained under \isovrm-VR and \isovrm-Q2 reach $51.5$ and $49.8$, compared with $49.0$ under VideoReward, the same order as their reward models' transfer accuracy on VBench-2.0 (Appendix~\ref{app:vbench2-generator-sec} gives every dimension).
Both intervention-trained reward models therefore yield generators that improve on the VideoReward-supervised one under human judgments and under VBench-2.0.

\section{Related work}
\label{sec:related}
\textbf{Video reward models and their evaluation.}
Image and video reward models learn scores from human choices between independently generated outputs, through scalar or multidimensional heads, decomposed questions, reasoning-based scoring, or pairwise judgments~\citep{pickscore,imagereward,hpsv2,videoreward,videoscore,visionreward,videoscore2,lift,unifiedreward}, and general vision-language models are also prompted as judges~\citep{llmjudge,mllmjudge}.
Reward models are evaluated by agreement with human preferences between generated videos in VideoGen-RewardBench, GenAI-Bench, and MJ-Bench-Video~\citep{videoreward,genaibench,mjvideo}, and generators by per-dimension protocols in VBench, VBench-2.0, and EvalCrafter~\citep{vbench,vbench2,evalcrafter}.
Even when an annotation names the dimension to judge, the two videos differ in other dimensions and in generator identity, so a model can succeed on correlated differences instead of the requested property, a shortcut documented in other domains~\citep{annotationartifacts,shortcutlearning}.
Our pairs differ in one property by construction.

\textbf{Constructed supervision.}
Synthetic data transforms scarce source data into forms a model learns from more readily~\citep{yang2024syntheticcontinuedpretraining}.
Image-quality assessment applies distortions of known severity but still collects human quality labels~\citep{tid2013}, attribute-specific constructions support question generation and latent preference models~\citep{alfa,prefpalette}, and comparisons between stronger and weaker generators give preferences without human labeling but only for overall quality~\citep{deltalearning}.
Our generator substitutions reuse that signal, and the other interventions extend the construction to one named property per video pair, with no human labeling.

\textbf{Temporal order.}
Frame order has long served as self-supervision, for recognizing forward playback, detecting shuffled frames, and sorting clips~\citep{seeingarrow,arrowoftime,shuffleandlearn,sortingsequences,cliporder}.
Video-language benchmarks find models weak on temporal understanding beyond static content~\citep{temporalbench,tempcompass}, and TempCompass, the closest construction to ours, builds conflicting videos that share static content and differ in one temporal property, by reversal or by spatial or temporal concatenation (Table~\ref{tab:temporal-contrast}).
These benchmarks test a single video against text, and \temporalshuffle{} poses the same contrast as a preference pair.

\section{Conclusion}

Controlled interventions make a reward model's perception testable one property at a time, and they show that current video reward models see frame content but not frame order, a failure that none of five existing benchmarks reveals.
Because the construction determines the preferred side, the resulting pairs need no human labels, and human validation is required only once per intervention.
Used as supervision, these pairs train reward models that transfer to human-labeled benchmarks and yield generators people prefer, with or without human preference data.
Any property that can be written as an intervention can therefore be measured in a reward model and taught to it, without collecting new human preferences.

\subsection*{Ethics Statement}

\bench{} contains model-generated videos and does not require collecting videos
of research participants. Human participation is limited to perceptibility validation and
preference judgments on generated clips. Generated content and learned reward
models may inherit biases, unsafe depictions, or representational gaps from the
source prompts and base generators. These artifacts are intended for research
on video reward modeling and should not be used as standalone measures of
safety, factuality, or human values.

% \subsection*{Reproducibility Statement}

% Sections~\ref{sec:method} and~\ref{sec:results} specify the intervention
% construction and evaluation protocol. Sections~\ref{sec:method} and~\ref{sec:training} report data splits, Section~\ref{sec:analysis} reports frame budgets, and the appendix gives training hyperparameters and per-dimension generator results. 

% \subsection*{Author Contributions}

% All authors contributed to conceptualization, methodology, experimental design,
% result interpretation, and manuscript revision. The first author led
% implementation, data construction, model training, evaluation, and initial
% manuscript preparation. All authors reviewed and approved the final manuscript.

\subsection*{Acknowledgments}
This research was developed in part with funding from the Defense Advanced Research Projects Agency's (DARPA) SciFy program (Agreement No. HR00112520300). The views expressed are those of the author and do not reflect the official policy or position of the Department of Defense or the U.S.~Government.
This material is based in part upon work supported by the Defense Advanced Research Projects Agency and the Air Force Research Laboratory, contract number(s): FA8650-23-C-7316. Any opinions, findings and conclusions, or recommendations expressed in this material are those of the author(s) and do not necessarily reflect the views of AFRL or DARPA.
This research was supported by Coefficient Giving, the University of Washington Population Health Initiative, Amazon Health, the UW+Amazon Science Hub, and the Meta AIM program.

% \clearpage
\bibliographystyle{assets/plainnat}
\bibliography{iclr2027_conference}

% \clearpage
\beginappendix
{\let\newpage\relax\newpage
\appendix

\section{Accuracy over decided pairs}
\label{app:ties}

Published reward-model benchmarks usually leave ties out of the denominator.
Table~\ref{tab:decided} reports accuracy over decided pairs with each model's tie rate, and Table~\ref{tab:main} counts ties as wrong.
Only VideoReward, VideoScore2, and LiFT keep their rank under both rules.
VisionReward moves from seventh to second, and PickScore moves from second to fifth.

\begin{table}[h]
\centering\scriptsize\setlength{\tabcolsep}{3pt}
\begin{tabular}{lrr|rrrrrrrr}
\toprule
\textbf{Model} & \textbf{Avg.} & \textbf{tie \%} &
\familyhead{{\renewcommand{\myfontsize}{\tiny}\temporalshuffle{}}} &
\familyhead{{\renewcommand{\myfontsize}{\tiny}\framehold{}}} &

\familyhead{{\renewcommand{\myfontsize}{\tiny}\motionregen{}}} &

\familyhead{{\renewcommand{\myfontsize}{\tiny}\attributeedit{}}} &

\familyhead{{\renewcommand{\myfontsize}{\tiny}\reducedsteps{}}} &

\familyhead{{\renewcommand{\myfontsize}{\tiny}\alignedweakgen{}}} &

\familyhead{{\renewcommand{\myfontsize}{\tiny}\weakgenfive{}}} &

\familyhead{{\renewcommand{\myfontsize}{\tiny}\weakgenone{}}} \\
\midrule
VideoReward & 73.6 & 0 & 61.0 & 71.9 & 84.9 & 74.6 & 78.7 & 74.0 & 78.2 & 65.2 \\
VisionReward & 71.3 & 21 & 47.4 & 56.9 & 67.0 & 70.0 & 83.4 & 76.8 & 89.2 & 80.1 \\
VideoScore v1 & 70.3 & 7 & 55.6 & 48.8 & 81.2 & 51.9 & 82.7 & 78.5 & 83.1 & 80.3 \\
UnifiedReward-Flex-9B & 69.8 & 2 & 48.5 & 51.8 & 55.2 & 79.1 & 77.6 & 71.9 & 91.1 & 83.5 \\
PickScore & 69.7 & 0 & 45.0 & 54.2 & 68.3 & 78.0 & 68.0 & 75.0 & 89.6 & 79.4 \\
UnifiedReward-2.0-8B & 69.5 & 2 & 53.1 & 54.6 & 57.4 & 72.4 & 77.4 & 68.6 & 90.5 & 81.9 \\
VideoScore v1.1 & 67.5 & 5 & 57.9 & 39.9 & 69.7 & 62.4 & 71.8 & 79.5 & 83.9 & 74.6 \\
VideoScore2 & 64.8 & 27 & 54.0 & 54.3 & 59.4 & 72.5 & 67.0 & 67.1 & 76.0 & 68.2 \\
LiFT & 50.7 & 61 & 43.0 & 43.4 & 67.2 & 52.3 & 48.0 & 57.0 & 40.4 & 54.0 \\
\bottomrule
\end{tabular}
\caption{Accuracy over decided pairs only, with the share of pairs on which each model expressed no preference.
Compare Table~\ref{tab:main}, which counts those pairs as wrong.}
\label{tab:decided}
\end{table}

\section{Additional validation}
\label{app:validation}

\subsection{Confinement has to be judged frame by frame}

Judging visual quality on whole videos mixes appearance with motion.
Human raters preferred the \motionregenname anchor on $17$ of $20$ decided video pairs but on none of $17$ corresponding frame pairs.
We therefore test confinement on single frames.

For each of $45$ clips, raters compare the anchor and \motionregenname{} frames at the index where the two differ most, excluding the keyframes, which are identical.
Raters judged $41$ pairs equal, and the other four split three to one ($p=0.63$, sign test).
All eight identical controls were judged equal.
The regenerated intervals use $30$ denoising steps against the anchor's $50$, and the frame comparison shows no loss of per-frame quality from it.

\subsection{Intervention strength sets difficulty}
\label{app:strength}

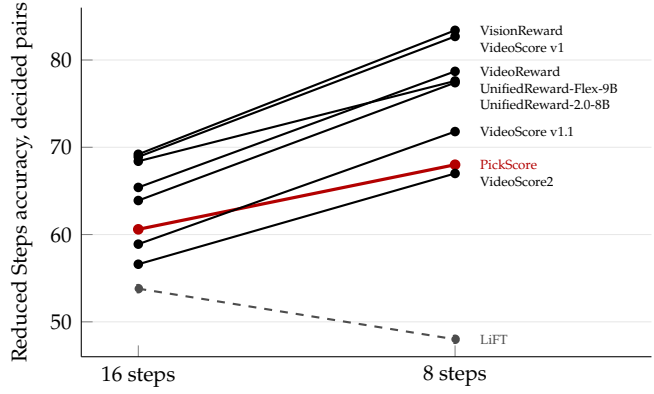
\begin{figure}[h]
\centering
\begin{tikzpicture}
\begin{axis}[
  width=0.55\linewidth, height=6.2cm,
  xmin=-0.18, xmax=1.62, ymin=46, ymax=86,
  xtick={0,1}, xticklabels={16 steps, 8 steps},
  ylabel={Reduced Steps accuracy, decided pairs},
  ymajorgrids, grid style={gray!18},
  axis x line*=bottom, axis y line*=left,
  label style={font=\footnotesize}, tick label style={font=\footnotesize},
  clip=false,
]
\addplot[thick, black, mark=*, mark size=1.4pt] coordinates {(0,65.4) (1,78.7)};
\node[font=\tiny, anchor=west] at (axis cs:1.05,78.7) {VideoReward};
\addplot[thick, black, mark=*, mark size=1.4pt] coordinates {(0,68.4) (1,77.6)};
\node[font=\tiny, anchor=west] at (axis cs:1.05,76.8) {UnifiedReward-Flex-9B};
\addplot[very thick, red!70!black, mark=*, mark size=1.4pt] coordinates {(0,60.6) (1,68.0)};
\node[font=\tiny, anchor=west, red!70!black] at (axis cs:1.05,68.0) {PickScore};
\addplot[thick, black, mark=*, mark size=1.4pt] coordinates {(0,63.9) (1,77.4)};
\node[font=\tiny, anchor=west] at (axis cs:1.05,74.9) {UnifiedReward-2.0-8B};
\addplot[thick, black, mark=*, mark size=1.4pt] coordinates {(0,68.9) (1,82.7)};
\node[font=\tiny, anchor=west] at (axis cs:1.05,81.5) {VideoScore v1};
\addplot[thick, black, mark=*, mark size=1.4pt] coordinates {(0,58.9) (1,71.8)};
\node[font=\tiny, anchor=west] at (axis cs:1.05,71.8) {VideoScore v1.1};
\addplot[thick, black, mark=*, mark size=1.4pt] coordinates {(0,69.2) (1,83.4)};
\node[font=\tiny, anchor=west] at (axis cs:1.05,83.4) {VisionReward};
\addplot[thick, black, mark=*, mark size=1.4pt] coordinates {(0,56.6) (1,67.0)};
\node[font=\tiny, anchor=west] at (axis cs:1.05,66.1) {VideoScore2};
\addplot[thick, gray!60!black, dashed, mark=*, mark size=1.4pt] coordinates {(0,53.8) (1,48.0)};
\node[font=\tiny, anchor=west, gray!60!black] at (axis cs:1.05,48.0) {LiFT};
\end{axis}
\end{tikzpicture}
\caption{Accuracy on decided pairs against the number of denoising steps on \reducedstepsname, with prompt, seed, anchor and every other sampling parameter held fixed.
Eight of the nine models gain as the budget falls from $16$ to $8$ denoising steps, by $7.4$ to $14.2$ points.
PickScore, in red, gains least.
LiFT, dashed, falls by $5.8$.
Tie rates are in Table~\ref{tab:decided}.}
\label{fig:steps}

\end{figure}

We render \reducedstepsname{} at $8$ and $16$ denoising steps with the prompt, seed, anchor, and every other sampling parameter fixed, and score all nine reward models on both over decided pairs (Figure~\ref{fig:steps}).

Eight models gain $7.4$ to $14.2$ points at the stronger degradation.
VideoReward passes UnifiedReward-Flex-9B, and VideoScore v1.1 passes PickScore.
PickScore, which scores frames independently, gains least.
LiFT falls from $53.8$ to $48.0$ and is the only model below chance at $8$ steps.
The spread across models widens from $53.8$--$69.2$ at $16$ steps to $48.0$--$83.4$ at $8$.
We release the $16$-step setting as an additional split outside the $8{,}000$ benchmark pairs.

On \reducedsteps{}, the strength of the intervention therefore sets its difficulty for reward models.

\section{Temporal Shuffle anchor selection}
\label{app:shuffle-motion}

Reordering is visible only in a clip that moves, so \temporalshuffle{} draws on the anchors with the most motion.
We measure motion as the mean absolute difference between sampled frames in 8-bit pixel values, averaged over pixels and color channels.
For the benchmark, motion is computed on every third frame, and the $1{,}000$ anchors with the most motion range from $18.3$ to $56.1$, against about $12$ for a typical clip.
The training corpus instead keeps every anchor above $25$, computed on every fourth frame, which admits $23\%$ of them.

\section{Tie handling}
\label{app:tiehandling}

Tie rates range from zero to $61\%$ and follow how finely each model scores.
PickScore and VideoReward each produce about $15{,}000$ distinct scores and rarely tie.
LiFT produces ten, assigns $0.9$ to $40\%$ of clips, and ties on $61\%$ of pairs.
VisionReward and VideoScore2 tie on $21\%$ and $27\%$.

Paired clips share a prompt and most of their content, which raises the tie rate further for coarse scores.
Independent draws from LiFT's score distribution would tie on $25.8\%$ of pairs, against the observed $61.3\%$.
We count a tie as wrong because it gives no ordering for a difference people perceive.

\section{Prompted judge protocol}
\label{app:judges}

Each judge answers with a digit from $0$ to $9$, and we rescale the expected digit under its token probabilities to $0$--$100$.
For Qwen3-VL the probabilities come from its logits.
For GPT-4.1 they come from the ten most likely tokens the API returns, which hold nearly all the probability of a single-digit answer.

Taking the decoded digit instead produces many ties.
On samples of $200$ pairs for Qwen3-VL and $60$ for GPT-4.1, Qwen3-VL ties on $47\%$--$54\%$ of pairs and GPT-4.1 on $17\%$--$20\%$.
The expected digit brings both below $1\%$.
Asking Qwen3-VL for two decimals between $0$ and $1$ still yields only nine distinct values and a $54\%$ tie rate.
We use the expected digit throughout, which changes the judges' average rank under the tie rule but not their frame-order results, so we compare judges on the frame-order columns.

\section{Frames sampled per model}
\label{app:frames}

Table~\ref{tab:frames} lists how many frames each model samples from a $5.1$-second clip, next to its accuracy on the two frame-reuse interventions.

\begin{table}[h]
\centering\small
\begin{tabular}{lrrr}
\toprule
\textbf{Model} & \textbf{Frames} & \temporalshuffle{} & \framehold{}\\
\midrule
VideoReward & 10 & \textbf{60.8} & \textbf{71.6} \\
UnifiedReward-2.0-8B & 8 & 52.6 & 54.3 \\
PickScore & 6 & \underline{45.0} & 54.2 \\
UnifiedReward-Flex-9B & 8 & \underline{47.7} & \underline{49.5} \\
VideoScore v1 & 48 & \underline{49.5} & \underline{39.8} \\
VideoScore2 & 10 & \underline{38.2} & \underline{35.9} \\
VideoScore v1.1 & 48 & 54.2 & \underline{34.9} \\
VisionReward & 24 & \underline{27.7} & \underline{31.7} \\
LiFT & 8 & \underline{11.9} & \underline{8.9} \\
\bottomrule
\end{tabular}
\caption{Frames sampled per clip against accuracy on the two interventions that reuse the anchor's frames.
Clips are $81$ frames at $16$ fps.
Underlined entries fall below chance.
Frame count orders neither column.
Accuracies count a tie as wrong, as in Table~\ref{tab:main}, and so run lower than in Table~\ref{tab:external}, which drops ties as VideoGen-RewardBench does.}
\label{tab:frames}
\end{table}

Each model has its own architecture, so this comparison cannot separate frame count from model, and Section~\ref{sec:framesweep} varies the count within each model instead.

\section{Reward model training}
\label{app:training-hparams}

Both released models use the same training configuration and differ only in initialization.
They train on $160{,}000$ pairs and validate on $4{,}000$, drawn equally from the eight interventions.
The training, validation, and benchmark prompt sets are disjoint.

\begin{table}[h]
\centering\small
\begin{tabular}{ll}
\toprule
\textbf{Setting} & \textbf{Value} \\
\midrule
Backbone & Qwen2-VL-2B-Instruct \\
Initialization & VideoReward, or the bare backbone \\
Reward head & scalar, read at three per-dimension marker tokens \\
Objective & Bradley-Terry with ties \\
LoRA & rank $64$, $\alpha$ $128$, dropout $0.05$ \\
LoRA exclusions & \texttt{lm\_head}, \texttt{rm\_head}, \texttt{embed\_tokens}, \texttt{visual} \\
Vision tower & trained, LoRA not applied \\
Learning rate & $2\times10^{-6}$ \\
Schedule & constant with warmup, warmup ratio $0.05$ \\
Weight decay & $0$ \\
Epochs & $3$ \\
Batch size & $1$ per device, no gradient accumulation \\
Hardware & $8$ GPUs on one node \\
Precision & bfloat16 \\
Seed & $42$ \\
Frames & $2$ fps, at most $200{,}704$ pixels per frame \\
\bottomrule
\end{tabular}
\caption{Training configuration for both released reward models.
}
\label{tab:training-hparams}
\end{table}

\section{The trained reward models on \bench{}}
\label{app:ours-iso}

Table~\ref{tab:ours-iso} reports the two trained reward models on \bench{}, with ties counted as wrong.
Training excludes every benchmark prompt but uses the same eight interventions, so this tests new prompts and not new interventions.

\begin{table}[h]
\centering\scriptsize\setlength{\tabcolsep}{2.5pt}
\begin{tabular}{lr|rrr|r|r|rrr}
\toprule
 & & \multicolumn{3}{c|}{\textbf{motion}} & \textbf{text} & \textbf{visual} & \multicolumn{3}{c}{\textbf{overall}} \\
\textbf{Model} & \textbf{Avg.} &
\familyhead{{\renewcommand{\myfontsize}{\tiny}\temporalshuffle{}}} &
\familyhead{{\renewcommand{\myfontsize}{\tiny}\framehold{}}} &

\familyhead{{\renewcommand{\myfontsize}{\tiny}\motionregen{}}} &

\familyhead{{\renewcommand{\myfontsize}{\tiny}\attributeedit{}}} &

\familyhead{{\renewcommand{\myfontsize}{\tiny}\reducedsteps{}}} &

\familyhead{{\renewcommand{\myfontsize}{\tiny}\alignedweakgen{}}} &

\familyhead{{\renewcommand{\myfontsize}{\tiny}\weakgenfive{}}} &

\familyhead{{\renewcommand{\myfontsize}{\tiny}\weakgenone{}}} \\
\midrule
VideoReward (2B), human preferences & 73.3 & 60.8 & 71.6 & 84.7 & 74.1 & 78.5 & 73.6 & 77.9 & 64.9 \\
\isovrm-Q2 & 96.3 & 84.9 & 99.8 & 100.0 & 86.8 & 100.0 & 99.9 & 100.0 & 99.3 \\
\isovrm-VR & 96.7 & 89.5 & 100.0 & 100.0 & 85.5 & 99.7 & 99.9 & 99.7 & 99.0 \\
\bottomrule
\end{tabular}
\caption{The two released reward models on \bench{}, ties counted as wrong, with VideoReward from Table~\ref{tab:main} for reference.}
\label{tab:ours-iso}
\end{table}

The two models average $23.0$ and $23.4$ points above VideoReward.
Both reach at least $99.0\%$ on six interventions, with the remaining errors concentrated on \temporalshuffle{} and \attributeedit{}.
VideoReward initialization raises \temporalshuffle{} accuracy from $84.9\%$ to $89.5\%$, while the Qwen2-VL initialization is slightly better on \attributeedit{} at $86.8\%$ versus $85.5\%$.
Their averages differ by $0.4$ points.
Transfer to benchmarks built from human labels is in Table~\ref{tab:transfer}.

\section{Generator identity in preference benchmarks}
\label{app:generator-identity}

Preference benchmarks pair videos from a fixed set of generators that differ widely in quality.
On VideoGen-RewardBench, the share of its comparisons that each generator wins ranges from $16.8\%$ for EasyAnimate v4 to $80.9\%$ for Kling 1.5.
To measure how much of each benchmark's labels this explains, we predict the preferred video of every pair from the identities of its two generators alone.
For each pair, the predictor chooses the generator that wins more often against the other in the remaining four folds of a five-fold split, and it never sees a video.

\begin{table}[h]
\centering\small
\begin{tabular}{lrrrr}
\toprule
\textbf{Benchmark} & \textbf{Generators} & \textbf{Generator identity} & \textbf{VideoReward} & \textbf{LiFT} \\
\midrule
VideoGen-RewardBench & $12$ & $77.4$ & $73.2$ & $56.8$ \\
GenAI-Bench          & $10$ & $74.1$ & $69.9$ & $50.7$ \\
VBench-2.0           & $4$  & $78.2$ & $50.6$ & $58.9$ \\
\bottomrule
\end{tabular}
\caption{Accuracy (\%) of predicting each benchmark's human preference from the identities of the two generators alone, by five-fold cross-validation, next to the reward models with the highest and lowest average on \bench{}.
Chance is $50$.
Every column uses the pairs of Table~\ref{tab:external}, which for VBench-2.0 are the $2{,}494$ decided pairs of its four dimensions.}
\label{tab:generator-only}
\end{table}

Generator identity alone predicts $74.1\%$ to $78.2\%$ of the labels (Table~\ref{tab:generator-only}), and on each benchmark it is more accurate than VideoReward, the reward model with the highest average on \bench{}, while LiFT, the one with the lowest average, is near chance on GenAI-Bench.
A reward model can therefore gain accuracy on these benchmarks by recognizing which generator made each clip, without judging the property the label refers to, and Appendix~\ref{app:shortcut-failure} shows that VideoReward does so on VBench-2.0.

\section{Generator ranking in VideoReward and in our reward models}
\label{app:shortcut-failure}

Table~\ref{tab:vbench-genrank} compares each VBench-2.0 generator's win rate under human labels with the win rate each model's scores imply.
It shows where the $7.2$-point gap between \isovrm-VR and VideoReward on VBench-2.0 in Table~\ref{tab:transfer} comes from.

\begin{table}[h]
\centering\small
\begin{tabular}{lrrrrrr}
\toprule
\textbf{Generator} & \textbf{In VideoReward's} & \textbf{VBench-2.0} &
\multicolumn{2}{c}{\textbf{VideoReward}} & \multicolumn{2}{c}{\textbf{\isovrm-VR}} \\
\cmidrule(lr){4-5}\cmidrule(lr){6-7}
 & \textbf{training data} & \textbf{people} & win & $\Delta$ & win & $\Delta$ \\
\midrule
Kling        & $59.3$ / $80.9$ & $64.6$ & $70.9$ & $+6.3$  & $81.2$ & $+16.6$ \\
Sora         & absent          & $61.9$ & $47.9$ & $-14.0$ & $70.2$ & $+8.3$  \\
HunyuanVideo & absent          & $47.6$ & $41.1$ & $-6.5$  & $42.9$ & $-4.7$  \\
CogVideo     & $44.8$          & $28.1$ & $40.9$ & $+12.8$ & $9.7$  & $-18.4$ \\
\bottomrule
\end{tabular}
\caption{Percentage of its comparisons each generator wins, by human count and as each model's scores imply, with each model's deviation from the human value.
The first column is that generator's standing in VideoGen-RewardBench, the data VideoReward was trained on.
Kling has two entries in VideoGen-RewardBench and VBench-2.0 uses a third version, so its training-set entry is the least certain of the four.}
\label{tab:vbench-genrank}
\end{table}

VideoReward rates both generators present in its training data above their human win rates and both absent generators below them.
It assigns CogVideo $40.9$, within $4$ points of its $44.8$ VideoGen-RewardBench win rate and $13$ points above its $28.1$ VBench-2.0 human win rate.
With four generators, this pattern arises by chance one time in eight.

Five of our interventions use the same generator on both sides and so carry no generator ranking.
The inherited ranking affects VideoReward on VBench-2.0, whose generators it partly never saw, and not on VideoGen-RewardBench, whose generators it saw in training.

On VideoGen-RewardBench, GenAI-Bench, and MJ-Bench-Video, dropping the generator substitutions costs $1.9$, $1.6$, and $0.5$ points, $69.6$, $66.3$, and $60.7$ against $71.5$, $67.9$, and $61.2$ for \isovrm-Q2, so the five same-generator interventions supply most of the transfer.

\section{Rank on overall quality does not predict rank on frame order}
\label{app:indep}

The three generator substitutions carry an overall label, like the independently generated pairs of published benchmarks.
Table~\ref{tab:indep} compares each model's mean over them with its \temporalshuffle{} accuracy on the same prompts, corpus, and scoring rule.

\begin{table}[h]
\centering\small
\begin{tabular}{lrrrr}
\toprule
& \multicolumn{2}{c}{\textbf{Overall}} & \multicolumn{2}{c}{\temporalshuffle{}} \\
\textbf{Model} & mean & rank & mean & rank \\
\midrule
PickScore             & \textbf{81.3} & 1 & \underline{45.0} & 8 \\
UnifiedReward-Flex-9B & 79.7 & 2 & \underline{47.7} & 7 \\
VideoScore v1         & 78.2 & 3 & \underline{49.5} & 6 \\
UnifiedReward-2.0-8B  & 77.6 & 4 & 52.6 & 5 \\
VideoScore v1.1       & 77.2 & 5 & 54.2 & 4 \\
GPT-4.1               & 77.0 & 6 & 58.9 & 2 \\
Qwen3-VL-8B           & 74.8 & 7 & 57.1 & 3 \\
VideoReward           & 72.1 & 8 & \textbf{60.8} & 1 \\
VisionReward          & 71.2 & 9 & \underline{27.7} & 10 \\
VideoScore2           & 52.5 & 10 & \underline{38.2} & 9 \\
LiFT                  & \underline{22.7} & 11 & \underline{11.9} & 11 \\
\bottomrule
\end{tabular}
\caption{Mean accuracy over the three generator-substitution interventions, which carry an overall label because a weaker generator degrades every dimension at once, against accuracy on \temporalshuffle{}.
Underlined entries are below chance.
LiFT is the only model to hold its rank, at the bottom of both.}
\label{tab:indep}
\end{table}

PickScore ranks first of eleven on generator-substitution pairs and eighth on \temporalshuffle{} at $45.0$.
VideoReward ranks eighth on generator-substitution pairs and first on \temporalshuffle{}.
Without LiFT, the Pearson correlation between the overall mean and accuracy is $r=+0.40$ for \temporalshuffle{} and $r=+0.31$ for \framehold{}, and neither differs from zero.
With LiFT, which is below chance on all eight interventions, they rise to $+0.79$ and $+0.72$.

The two best models on generator-substitution pairs do not exceed chance on frame order.

\section{Frame counts the sweep cannot reach}
\label{app:framesweep}

Figure~\ref{fig:framesweep} omits three sets of points.
LiFT stops at $16$ frames because its VILA1.5-13B backbone has $4{,}096$ positions, and more frames push the prompt out of context so that every clip gets the top score.
VideoScore2 stops at $16$ because with more frames it fails to finish on half the clips, repeating a closing tag until its token limit without producing a score.
VideoReward stops at $32$ because $48$ frames exceed the memory of an $80$ GB GPU at batch size one.

\section{Caption ranking on temporal-understanding benchmarks}
\label{app:temporal}

A reward model gives one score per video and text, so we use the TemporalBench~\citep{temporalbench} and TempCompass~\citep{tempcompass} settings in which it picks among complete captions for a fixed video.
TemporalBench poses binary caption questions on short videos and reports Multiple Binary Accuracy, and TempCompass poses binary caption selection and reports caption-matching accuracy.

We score each caption with the model's text-alignment head, or its only output for a single-score model, and select the unique maximum, counting exact ties as wrong.
For UnifiedReward we use the continuous alignment score of its pointwise video prompt.

Table~\ref{tab:temporal-contrast} sets the three designs side by side.

\begin{table}[h]
\centering\scriptsize
\setlength{\tabcolsep}{3pt}
\resizebox{\linewidth}{!}{%
\begin{tabular}{llll}
\toprule
\textbf{Design property} &  \bench{} & \textbf{TemporalBench} &
\textbf{TempCompass} \\
\midrule
Original target & \shortstack[l]{Video-generation\\reward models} &
\shortstack[l]{Video text-generation\\and embedding models} & Video and image language models \\
What the model ranks & \shortstack[l]{One prompt, rank\\two videos} &
\shortstack[l]{One video, rank\\binary captions} &
\shortstack[l]{One video, rank\\two captions} \\
What changes & One named video property & A caption detail & The candidate
caption \\
Human role & \shortstack[l]{Validate each intervention,\\with no per-item label} &
\shortstack[l]{Write and refine positive captions,\\then review generated negatives} &
\shortstack[l]{Annotate video metadata,\\then review generated instructions} \\
Frame-order isolation & \shortstack[l]{Same prompt and frames,\\reordered} &
\shortstack[l]{None, since the evaluated\\video is fixed} &
\shortstack[l]{None, since the evaluated\\video is fixed} \\
\bottomrule
\end{tabular}}
\caption{Differences among the benchmark designs.
The TemporalBench and TempCompass columns describe the settings evaluated here.}
\label{tab:temporal-contrast}
\end{table}

TemporalBench counts a video correct only when every one of its binary questions is answered correctly, the group-level score its paper treats as primary, and we report it over $2{,}179$ videos.

TempCompass contains $1{,}503$ binary choices between complete captions.
Its other three tasks do not fit a single score, since multiple-choice answers are fragments such as ``up'' or ``forward'' that need the question to make sense, yes/no questions need a calibrated threshold, and caption generation needs a language model.

\begin{table}[h]
\centering\small
\setlength{\tabcolsep}{7pt}
\begin{tabular}{lrr}
\toprule
\textbf{Model} & \textbf{TemporalBench} & \textbf{TempCompass} \\
\midrule
VideoReward & $\mathbf{21.3}$ & $69.5$ \\
UnifiedReward-Flex-9B & $13.3$ & $63.3$ \\
PickScore & $13.5$ & $61.0$ \\
UnifiedReward-2.0-8B & $20.9$ & $68.3$ \\
VideoScore v1 & $7.0$ & $40.1$ \\
VideoScore v1.1 & $6.9$ & $43.0$ \\
VisionReward & $2.7$ & $47.4$ \\
VideoScore2 & $12.2$ & $57.1$ \\
LiFT & $2.7$ & $14.1$ \\
Ours, VideoReward init & $18.4$ & $\mathbf{69.7}$ \\
Ours, Qwen2-VL init & $18.1$ & $66.9$ \\
\bottomrule

\end{tabular}
\caption{Accuracy on the caption-ranking settings of TemporalBench and TempCompass.
All values are percentages.
Bold marks the best result in each column.}
\label{tab:temporal}
\end{table}

VideoReward has the highest TemporalBench score, $21.3$, and \isovrm-VR, initialized from it, has the highest TempCompass score, $69.7$, with $18.4$ on TemporalBench.

\section{Published benchmarks against \bench{}}
\label{app:external}

Figure~\ref{fig:temporal_shuffle} plots \temporalshuffle{} accuracy against each published benchmark, Table~\ref{tab:external} gives the values, and Table~\ref{tab:rhofamily} gives the rank correlation for every intervention.

\begin{table}[h]
\centering\scriptsize\setlength{\tabcolsep}{3pt}
\resizebox{\linewidth}{!}{%
\begin{tabular}{lrrrrr|rr}
\toprule
 & \multicolumn{5}{c|}{\textbf{Published Benchmarks}} & \multicolumn{2}{c}{\textbf{\bench{}}} \\
\textbf{Model} & VideoGen-RewardBench & GenAI-Bench & VBench-2.0 & TemporalBench & TempCompass & Temporal Shuffle & Frame Hold \\
\midrule
UnifiedReward-2.0-8B & $\mathbf{77.5}$ & $80.5$ & $63.3$ & $20.9$ & $68.3$ & $53.1$ & $54.6$ \\
UnifiedReward-Flex-9B & $74.5$ & $\mathbf{80.8}$ & $63.4$ & $13.3$ & $63.3$ & \underline{$48.5$} & $51.8$ \\
VideoReward & $73.2$ & $69.9$ & $50.6$ & $\mathbf{21.3}$ & $\mathbf{69.5}$ & $61.0$ & $71.9$ \\
VisionReward & $69.0$ & $73.8$ & $63.3$ & \underline{$2.7$} & \underline{$47.4$} & \underline{$47.4$} & $56.9$ \\
VideoScore2 & $67.7$ & $68.5$ & $53.5$ & $12.2$ & $57.1$ & $54.0$ & $54.3$ \\
PickScore & $62.6$ & $74.1$ & $57.4$ & $13.5$ & $61.0$ & \underline{$45.0$} & $54.2$ \\
VideoScore v1 & $61.1$ & $70.4$ & $\mathbf{66.2}$ & \underline{$7.0$} & \underline{$40.1$} & $55.6$ & \underline{$48.8$} \\
VideoScore v1.1 & $57.4$ & $71.5$ & $\mathbf{66.2}$ & \underline{$6.9$} & \underline{$43.0$} & $57.9$ & \underline{$39.9$} \\
LiFT & $56.8$ & $50.7$ & $58.9$ & \underline{$2.7$} & \underline{$14.1$} & \underline{$43.0$} & \underline{$43.4$} \\
\midrule
\emph{$\rho$ with Temporal Shuffle} & $+0.15$ & $-0.15$ & $+0.09$ & $+0.37$ & $+0.28$ & \multicolumn{2}{c}{} \\
\emph{$\rho$ with Frame Hold} & $+0.73$ & $+0.13$ & $-0.61$ & $+0.52$ & $+0.72$ & \multicolumn{2}{c}{} \\
\bottomrule
\end{tabular}}
\caption{Accuracy on five published benchmarks against
accuracy on the two \bench{} interventions that reuse the anchor's frames. All
columns use each benchmark's reported metric, and the \bench{} columns use the
decided-item protocol of Table~\ref{tab:decided}. Underlined entries fall below
the corresponding random baseline. The last two rows give
the Spearman rank correlation down each benchmark's column. Bold marks the best
model in each published-benchmark column.}
\label{tab:external}
\end{table}

\begin{table}[h]
\centering\small\setlength{\tabcolsep}{5pt}
\begin{tabular}{lrrrrrrrr}
\toprule
\textbf{Benchmark} &
\familyhead{{\renewcommand{\myfontsize}{\tiny}\temporalshuffle{}}} &
\familyhead{{\renewcommand{\myfontsize}{\tiny}\framehold{}}} &

\familyhead{{\renewcommand{\myfontsize}{\tiny}\motionregen{}}} &

\familyhead{{\renewcommand{\myfontsize}{\tiny}\attributeedit{}}} &

\familyhead{{\renewcommand{\myfontsize}{\tiny}\reducedsteps{}}} &

\familyhead{{\renewcommand{\myfontsize}{\tiny}\alignedweakgen{}}} &

\familyhead{{\renewcommand{\myfontsize}{\tiny}\weakgenfive{}}} &

\familyhead{{\renewcommand{\myfontsize}{\tiny}\weakgenone{}}} \\
\midrule
VideoGen-RewardBench & $+0.15$ & $\mathbf{+0.73}$ & $-0.50$ & $+0.63$ & $+0.43$ & $-0.20$ & $+0.62$ & $+0.57$ \\
GenAI-Bench & $-0.15$ & $+0.13$ & $-0.48$ & $+0.50$ & $+0.37$ & $+0.27$ & $\mathbf{+1.00}$ & $\mathbf{+0.87}$ \\
VBench-2.0 & $+0.09$ & $-0.61$ & $-0.08$ & $-0.46$ & $+0.34$ & $+0.56$ & $+0.39$ & $+0.60$ \\
TemporalBench & $+0.37$ & $+0.52$ & $+0.03$ & $+0.67$ & $+0.07$ & $-0.15$ & $+0.34$ & $+0.19$ \\
TempCompass & $+0.28$ & $\mathbf{+0.72}$ & $-0.20$ & $\mathbf{+0.80}$ & $+0.22$ & $-0.17$ & $+0.48$ & $+0.27$ \\
\bottomrule
\end{tabular}
\caption{Spearman rank correlation between accuracy on each published
benchmark and accuracy on each \bench{} intervention, over the 9 reward models
common to all columns. Each benchmark uses its reported metric. Bold marks a
correlation an exact permutation test separates from zero at
$p < 0.05$. VBench-2.0 correlates negatively with Frame Hold, at $-0.61$ and
$-0.80$ without LiFT, and TemporalBench's largest correlation is $+0.67$, with
Attribute Edit.}
\label{tab:rhofamily}
\end{table}

\section{Every intervention against the human ceiling}
\label{app:gradient}

Figure~\ref{fig:gradient} orders the interventions by median reward-model accuracy, and the two frame-reuse interventions come last.

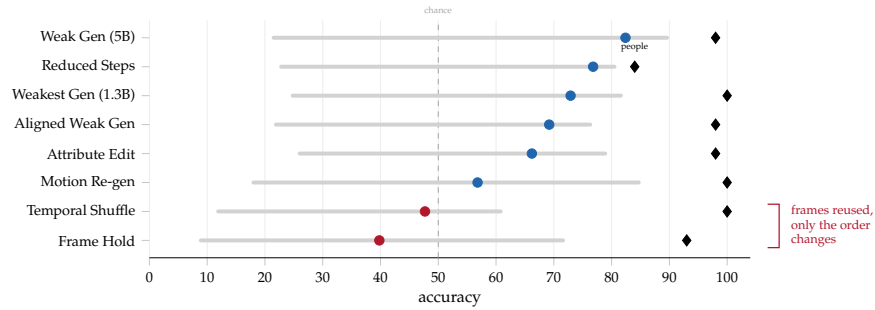
\begin{figure}[t]
\centering
\resizebox{0.7\linewidth}{!}{%
\begin{tikzpicture}
\begin{axis}[
  width=0.80\linewidth, height=6.2cm,
  xmin=0, xmax=104, xlabel={accuracy},
  ytick={1,...,8},
  yticklabels={%
    Frame Hold, Temporal Shuffle, Motion Re-gen, Attribute Edit,
    Aligned Weak Gen, Weakest Gen (1.3B), Reduced Steps, Weak Gen (5B)},
  ymin=0.4, ymax=8.6,
  xmajorgrids, grid style={gray!14},
  axis x line*=bottom, axis y line*=left, y axis line style={draw=none},
  tick align=outside, tick style={gray!45},
  label style={font=\small}, tick label style={font=\footnotesize},
  clip=false,
]
\addplot[dashed, gray!70] coordinates {(50,0.4) (50,8.6)};
\node[font=\tiny, gray!70, anchor=south] at (axis cs:50,8.62) {chance};

\draw[gray!35, line width=2.2pt, line cap=round] (axis cs:8.9,1) -- (axis cs:71.6,1);
\draw[gray!35, line width=2.2pt, line cap=round] (axis cs:11.9,2) -- (axis cs:60.8,2);
\draw[gray!35, line width=2.2pt, line cap=round] (axis cs:18.0,3) -- (axis cs:84.7,3);
\draw[gray!35, line width=2.2pt, line cap=round] (axis cs:26.0,4) -- (axis cs:78.9,4);
\draw[gray!35, line width=2.2pt, line cap=round] (axis cs:21.9,5) -- (axis cs:76.3,5);
\draw[gray!35, line width=2.2pt, line cap=round] (axis cs:24.8,6) -- (axis cs:81.6,6);
\draw[gray!35, line width=2.2pt, line cap=round] (axis cs:22.8,7) -- (axis cs:80.5,7);
\draw[gray!35, line width=2.2pt, line cap=round] (axis cs:21.5,8) -- (axis cs:89.6,8);

\addplot[only marks, mark=*, mark size=2.6pt, tbllo] coordinates {(39.8,1) (47.7,2)};
\addplot[only marks, mark=*, mark size=2.6pt, tblhi]
  coordinates {(56.8,3) (66.2,4) (69.2,5) (72.9,6) (76.8,7) (82.4,8)};

\addplot[only marks, mark=diamond*, mark size=3.1pt, black] coordinates {
  (93,1) (100,2) (100,3) (98,4) (98,5) (100,6) (84,7) (98,8)};
\node[font=\tiny, anchor=south, black] at (axis cs:84,7.3) {people};

\draw[tbllo, line width=0.7pt] (axis cs:107,0.75) -- (axis cs:109,0.75)
  -- (axis cs:109,2.25) -- (axis cs:107,2.25);
\node[font=\scriptsize, anchor=west, tbllo, align=left]
  at (axis cs:110,1.5) {frames reused,\\only the order\\changes};
\end{axis}
\end{tikzpicture}
}
\caption{Interventions ordered by median reward-model accuracy.
Each bar spans the nine reward models from worst to best, the dot marks their median, blue above chance and red below, and the diamond marks human agreement on the same $45$ pairs per intervention, scored the same way, which stays between $84$ and $100$.
On the two frame-reuse interventions the median model reaches $39.8$ and $47.7$, against $93$ and $100$ for people.
People score above the best reward model on every intervention, by more than the error of the $45$ rated pairs on all but \reducedsteps{} ($84$ against $80.5$) and \weakgenfive{} ($98$ against $89.6$).}
\label{fig:gradient}
\end{figure}

\section{Human preference comparisons for the trained generators}
\label{app:grpo-human}

Table~\ref{tab:grpo-human} reports three annotation passes over the pairwise comparisons among the three generators after $300$ training steps.
Each pass contains all $240$ comparisons, with the same $80$ held-out prompts used for each model pair.
Item order and side assignment are randomized independently.

\begin{table}[h]
\centering\small\setlength{\tabcolsep}{5pt}
\resizebox{0.5\linewidth}{!}{%
\begin{tabular}{lrrrr}
\toprule
\textbf{Preferred generator} & \textbf{Win Rate} & \textbf{Decisions} &
\textbf{Agreement} & $\boldsymbol{\kappa}$ \\
\midrule
\isovrm-Q2 over VideoReward & $56.2$ & $240$ & $63.3$ & $0.26$ \\
\isovrm-VR over VideoReward & $57.1$ & $240$ & $59.2$ & $0.17$ \\
\isovrm-Q2 over \isovrm-VR & $54.6$ & $240$ & $55.8$ & $0.11$ \\
\bottomrule

\end{tabular}
}\vspace{-5mm}
\caption{Blinded human preference comparisons on $80$ prompts.
Win rate pools the three annotation passes.
Agreement is the mean exact agreement over the three pairs of passes on the same comparison, and the final column gives Fleiss' $\kappa$ across the three passes.}
\label{tab:grpo-human}
\end{table}

Pooled over passes, both generators trained under our reward models win more comparisons than the VideoReward-supervised one.
Raters agree most on \isovrm-Q2 against VideoReward, $63.3\%$ with $\kappa = 0.26$, and least on \isovrm-Q2 against \isovrm-VR, $55.8\%$ with $\kappa = 0.11$, the comparison closest to even.
Over all $240$ comparisons, mean pairwise agreement is $59.4\%$ and Fleiss' $\kappa$ is $0.18$.
Those two generators also differ by only $1.7$ points on VBench-2.0.

Each pass is nearly transitive.
A pass sees the same three videos for each prompt, one per generator, and its three choices form a cycle on $5$ of the $240$ prompt and pass pairs, against $25\%$ for random choices (Table~\ref{tab:grpo-transitivity}).
Raters therefore rank the three generators consistently within a pass, and the low agreement comes from different rankings across passes.

\begin{table}[h]
\centering\small\setlength{\tabcolsep}{5pt}
\resizebox{0.4\linewidth}{!}{%
\begin{tabular}{lrrr}
\toprule
& \textbf{Cyclic Prompts} & \textbf{Prompts} & \textbf{Cyclic (\%)} \\
\midrule
Pass 1 & $1$ & $80$ & $1.2$ \\
Pass 2 & $4$ & $80$ & $5.0$ \\
Pass 3 & $0$ & $80$ & $0.0$ \\
\midrule
All passes & $5$ & $240$ & $2.1$ \\
Majority of passes & $2$ & $80$ & $2.5$ \\
\bottomrule

\end{tabular}
}\vspace{-5mm}
\caption{Transitivity of the human comparisons.
A prompt is cyclic for a pass when its three pairwise choices there form a cycle, such as \isovrm-Q2 over \isovrm-VR, \isovrm-VR over VideoReward, and VideoReward over \isovrm-Q2.
Random choices would give a cycle on $25\%$ of prompts.
The last row applies the majority of the three passes to each comparison.}
\label{tab:grpo-transitivity}
\end{table}

\section{VBench-2.0 by dimension for the trained generators}
\label{app:vbench2-generator-sec}

Table~\ref{tab:vbench2-generator} breaks the aggregate VBench-2.0 results into all eighteen dimensions.
Within each dimension we score every generator on the same number of videos per prompt, keeping the smaller count where two differ, since duplicate scoring across evaluation runs and four extra dimensions otherwise give unequal counts.
Dynamic\_Spatial\_Relationship, for example, has $74$ and $128$ videos over the same $69$ prompts.
Diversity uses the twenty generations per prompt it requires, which are equal by construction.

\begin{table}[H]
\centering\small\setlength{\tabcolsep}{4pt}
\resizebox{0.6\linewidth}{!}{%
\begin{tabular}{lrrrrr}
\toprule
\textbf{Dimension} & \textbf{Videos} & \textbf{Pretrained} & \textbf{VideoReward} & \textbf{\isovrm-Q2} & \textbf{\isovrm-VR} \\
\midrule
Camera\_Motion & $108$ & $15.7$ & $7.4$ & $\mathbf{16.7}$ & $10.2$ \\
Complex\_Landscape & $30$ & $14.7$ & $\mathbf{16.7}$ & $14.7$ & $16.0$ \\
Complex\_Plot & $61$ & $13.5$ & $\mathbf{14.5}$ & $11.7$ & $11.1$ \\
Composition & $53$ & $\mathbf{49.1}$ & $47.2$ & $43.0$ & $48.2$ \\
Diversity & $10$ & $47.6$ & $64.3$ & $60.9$ & $\mathbf{81.0}$ \\
Dynamic\_Attribute & $94$ & $\mathbf{12.8}$ & $11.7$ & $7.4$ & $8.5$ \\
Dynamic\_Spatial\_Relationship & $69$ & $27.5$ & $\mathbf{47.8}$ & $36.2$ & $30.4$ \\
Human\_Anatomy & $264$ & $84.2$ & $86.7$ & $\mathbf{91.3}$ & $87.7$ \\
Human\_Clothes & $191$ & $91.7$ & $97.8$ & $97.8$ & $\mathbf{100.0}$ \\
Human\_Identity & $128$ & $72.5$ & $76.9$ & $\mathbf{77.9}$ & $77.1$ \\
Human\_Interaction & $129$ & $56.6$ & $62.0$ & $55.8$ & $\mathbf{63.6}$ \\
Instance\_Preservation & $62$ & $\mathbf{91.9}$ & $\mathbf{91.9}$ & $\mathbf{91.9}$ & $\mathbf{91.9}$ \\
Material & $50$ & $\mathbf{58.3}$ & $51.7$ & $54.2$ & $56.0$ \\
Mechanics & $57$ & $\mathbf{64.4}$ & $54.8$ & $52.3$ & $55.3$ \\
Motion\_Order\_Understanding & $181$ & $\mathbf{19.9}$ & $17.1$ & $17.7$ & $13.8$ \\
Motion\_Rationality & $58$ & $\mathbf{44.8}$ & $43.1$ & $37.9$ & $41.4$ \\
Multi-View\_Consistency & $97$ & $30.7$ & $33.4$ & $\mathbf{70.3}$ & $70.0$ \\
Thermotics & $50$ & $55.3$ & $56.2$ & $59.1$ & $\mathbf{65.2}$ \\
\midrule
Mean & & $47.3$ & $49.0$ & $49.8$ & $51.5$ \\
\bottomrule
\end{tabular}
}\vspace{-2mm}
\caption{Every VBench-2.0 dimension for the four generators, in percent.
\textbf{Videos} is the count each is scored on, held equal across all four dimension by dimension.
Bold marks the best in a row.
Against the VideoReward-supervised run, the \isovrm-VR run leads on 11 dimensions and trails on 6; the median difference is $+0.8$ and a bootstrap over dimensions puts the mean difference in $[-1.9, +7.8]$.
}
\label{tab:vbench2-generator}
\end{table}

Fourteen of eighteen dimensions separate the two trained runs by less than $5$ points.
Removing Multi-View\_Consistency alone, or that dimension and Diversity, reverses the sign of the \isovrm-VR margin over the VideoReward-supervised run.

The aggregate weights every dimension equally, so Instance\_Preservation, at $91.9$ for all four generators, and Complex\_Landscape, spanning $2.0$ points, count as much as Multi-View\_Consistency, spanning $39.6$, and a dimension scored on $10$ videos counts as much as one scored on $264$.

\section{Generator training}
\label{app:grpo-hparams}

Each run draws from $12{,}303$ training prompts and samples eight videos per prompt, with advantages computed within each group.
Videos contain $121$ frames at $704\times1280$ and use $30$ sampling steps.
The reward weights visual and motion quality by $0.5$ and text alignment by $1.0$.
A KL penalty keeps the policy near a reference policy that is updated every $10$ steps.
Each run uses $128$ H200 GPUs and validates every $50$ steps on $512$ held-out prompts, with one video per prompt at $50$ denoising steps.
}

\end{document}